\pdfoutput=1

\documentclass[11pt]{article}

\usepackage[preprint]{acl}

\usepackage{times}
\usepackage{latexsym}
\usepackage{graphicx}

\usepackage{adjustbox}
\usepackage{etoolbox}

\usepackage{fourier} 
\usepackage{array}
\usepackage{makecell}
\usepackage{float}
\usepackage[]{hyperref}

\newcolumntype{P}[1]{>{\centering\arraybackslash}p{#1}}
\usepackage{tikz}
\usetikzlibrary{shapes,arrows}

\usetikzlibrary{shapes.geometric, arrows, positioning}
\usepackage{mdframed}
\usepackage{lipsum}
\usepackage{multirow}

\usepackage{array}
\usepackage{enumitem}
\usepackage{booktabs}
\usepackage{xurl}

\newmdtheoremenv{theo}{Theorem}
\usepackage{tabularx}

\usepackage{todonotes}

\usepackage[most]{tcolorbox}
\usepackage[T1]{fontenc}

\usepackage[utf8]{inputenc}

\usepackage{microtype}

\usepackage{inconsolata}
\usepackage{xspace}
\usepackage{graphicx}

\usepackage{listings}
\definecolor{lbcolor}{rgb}{0.95,0.95,0.95}
\newcommand{\myhl}[1]{\textcolor{red}{\textit{#1}}}

\newcommand{\papername}{TN-Eval\xspace}
\newcommand{\humaneval}{TN\textsuperscript{H}-Eval\xspace}
\newcommand{\autoeval}{TN\textsuperscript{A}-Eval\xspace}

\usepackage{acronym}
\acrodef{IAA}{Inter-Annotator Agreement}

\title{\papername: Rubric and Evaluation Protocols for Measuring the Quality of Behavioral Therapy Notes}

\author{
    \textbf{Raj Sanjay Shah\textsuperscript{1}\thanks{Work done during internship at Amazon}, Lei Xu\textsuperscript{2}, Qianchu Liu\textsuperscript{2}, Jon Burnsky\textsuperscript{2},}\\
    \textbf{Drew Bertagnolli\textsuperscript{3}, Chaitanya Shivade\textsuperscript{2}}\\
    \textsuperscript{1} Georgia Institute of Technology, 
    \textsuperscript{2} AWS AI Labs, 
    \textsuperscript{3} OneMedical \\
    \texttt{rajsanjayshah@gatech.edu}\\
    \texttt{\{ leixx, liufqian, jburnsky, shivadc \} @ amazon.com}\\
    \texttt{abertagnolli@onemedical.com} \\
}

\begin{document}
\maketitle

\begin{abstract}

Behavioral therapy notes are important for both legal compliance and patient care. Unlike progress notes in physical health, quality standards for behavioral therapy notes remain underdeveloped. 
To address this gap, we collaborated with licensed therapists to design a comprehensive rubric for evaluating therapy notes across key dimensions: \textit{completeness}, \textit{conciseness}, and \textit{faithfulness}. Further, we extend a public dataset of behavioral health conversations with therapist-written notes and LLM-generated notes, and apply our evaluation framework to measure their quality. We find that: (1) A rubric-based manual evaluation protocol offers more reliable and interpretable results than traditional Likert-scale annotations. (2) LLMs can mimic human evaluators in assessing completeness and conciseness but struggle with faithfulness. (3) Therapist-written notes often lack completeness and conciseness, while LLM-generated notes contain hallucination. Surprisingly, in a blind test, therapists prefer and judge LLM-generated notes to be superior to therapist-written notes. 
\end{abstract}

\section{Introduction}

Automated medical note generation using large language models (LLMs) has the potential to enhance clinicians' efficiency by reducing the time spent on electronic health records, allowing them to focus more on patient care. However, applying LLMs to behavioral health notes presents unique challenges~\citep{hua2024largelanguagemodelsmental}. 
In therapy, the conversation itself is the treatment; therefore, techniques like motivational interviewing used in a session may not be explicitly stated. Furthermore, sessions cover various topics, making it crucial to discern significant details from less relevant information.
Given the high-stakes nature of behavioral health, using LLMs to generate notes must be rigorously evaluated to ensure they capture key information at an appropriate level of detail.

Evaluating the quality of talk therapy notes, however, is not straightforward. Traditionally, human evaluation has been the primary method for assessing their quality, making it resource intensive and costly. 
Moreover, a lack of standardized reference notes and the limited literature on what constitutes an effective behavioral health therapy note further complicates the evaluation process. Therapists and healthcare providers often have their own styles and preferences, leading to subjective assessments and considerable variation. Without clear standards and evaluation protocols, it becomes difficult to determine the quality of LLM-generated therapy notes.

In this work, we focus on the SOAP (Subjective, Objective, Assessment, Plan) format of therapy notes and propose an evaluation framework for notes (\papername). The framework includes (1) a comprehensive, fine-grained, section-wise rubric that outlines the key components and characteristics of a therapy note and (2) both human and automatic evaluation protocols. The rubric, which we co-designed with 5 licensed therapists, details the relevant items for each of the four SOAP sections and their respective levels of importance \textbf{(Section~\ref{sec:rubric})}.  We then design a human evaluation protocol -- \humaneval\space -- in which 9 licensed behavioral health therapists from diverse backgrounds assess notes along three dimensions: completeness, conciseness, and faithfulness \textbf{(Section~\ref{sec:method_human})}. The completeness and conciseness are scored with reference to the rubric to improve the consistency of the evaluation, while faithfulness is evaluated at the sentence level with source attribution~\citep{rashkin2023measuring}. Finally, we explore the potential of LLMs to emulate expert evaluations, introducing an automatic evaluation protocol called \autoeval \textbf{(Section~\ref{sec:method_auto})}.

Our experimental results show that our proposed human evaluation protocol -- \humaneval achieves higher \ac{IAA} compared to conventional Likert-scale human evaluation, making it more reliable. 
We additionally show that using the automatic evaluation protocol -- \autoeval, we can achieve a better correlation with \humaneval on completeness and conciseness evaluation when compared to N-gram-based metrics like ROUGE~\cite{lin-2004-rouge} or conventional LLM-as-a-Judge~\cite{zheng2023judging}, making it a quick and cost-effective solution for evaluation. 
When compared to expert-written notes, we find that LLM-generated notes achieve around 10\% higher scores in completeness and conciseness but show relatively lower faithfulness.


\paragraph{Deployment considerations:} Our \autoeval is a \textit{deployable} and \textit{scalable} framework for assessing therapy notes with fine-grained, human-like judgments, which is designed by domain experts and has been evaluated on available datasets. Integrating this evaluation into clinical workflows and EHR systems enables: (1) Automated review that flags low-quality notes; (2) Automated scoring systems that assist therapists in refining notes before submission, reducing post-session documentation workload; and (3) Cost-effective, scalable quality assessments in standardized documentation practices. Refer appendix section \ref{subsec:intergration-proposal} for workflow integration suggestions.


\section{Related Work}
\textbf{AI in Mental Health Care:}
Recently, interest in using LLMs for mental health care has grown~\citep{Greer2019UseOT, peng2020exploring,srivastava2022counseling,luo2025mental}, with research focusing on three main directions. First, to classify therapeutic methods used by clinicians, assess the effectiveness of treatments, and predict the quality of service~\citep{Saha_Sharma_2020, chikersal2020understanding, liu-etal-2021-towards, shah2022modeling}. Second, virtual counselors emulate human behavior in chatbot-like environments~\citep{shen2020counseling, oneil-etal-2023-automatic}, but ethical and legal concerns~\citep{woodnutt2024could, stade2024large} have shifted research toward augmenting therapists with suggestions to enhance their responses~\citep{saha2022towards, sharma2023human}. Third, AI tools train novice counselors by providing automatic feedback~\citep{chaszczewicz-etal-2024-multi, lin-etal-2024-imbue} and simulating client personas for role-play~\citep{chen2023llmempowered, stapleton2023seeing, wang2024patient, louie2024roleplay}. Despite growing interest in AI for mental health support, LLMs for behavioral therapy note generation remain underexplored.

\noindent \textbf{Automated clinical note generation:} improves clinician efficiency~\citep{joshi-etal-2020-dr}, with research primarily focused on physical health using role-play or anonymized conversations and human-written notes~\citep{papadopoulos-korfiatis-etal-2022-primock57, ben-abacha-etal-2023-empirical, yim2023aci}. Early work fine-tuned lightweight transformer models~\citep{sharma-etal-2023-team, michalopoulos-etal-2022-medicalsum, milintsevich-agarwal-2023-calvados, yuan2024continued}, while recent studies explore LLM prompting for summarization~\citep{ben-abacha-etal-2023-empirical, mathur-etal-2023-summqa}.

\noindent \textbf{Automatic evaluations for summarization:} Reference-based metrics like ROUGE~\citep{lin-2004-rouge}, BLEU~\citep{papineni-etal-2002-bleu}, and BERTScore~\citep{zhang2019bertscore} measure lexical similarity between generated and reference summaries while fact-checking evaluators assess faithfulness to the source~\citep{honovich-etal-2022-true-evaluating, zha-etal-2023-alignscore, laban-etal-2022-summac}. Recently, LLMs have been increasingly used as evaluators, known as LLM-as-a-Judge~\citep{zheng2023judging, wang-etal-2023-chatgpt}. However, these methods are usually developed for general text summarization tasks and do not account for the challenges of therapy notes, where obtaining high-quality reference summaries is complex, and evaluations require substantial domain knowledge. Our \autoeval builds upon existing methodologies, adapting them to address these nuances.

\begin{figure*}[htb]
    \centering
 \includegraphics[width=.9\textwidth,trim=3 0 3 0,clip]{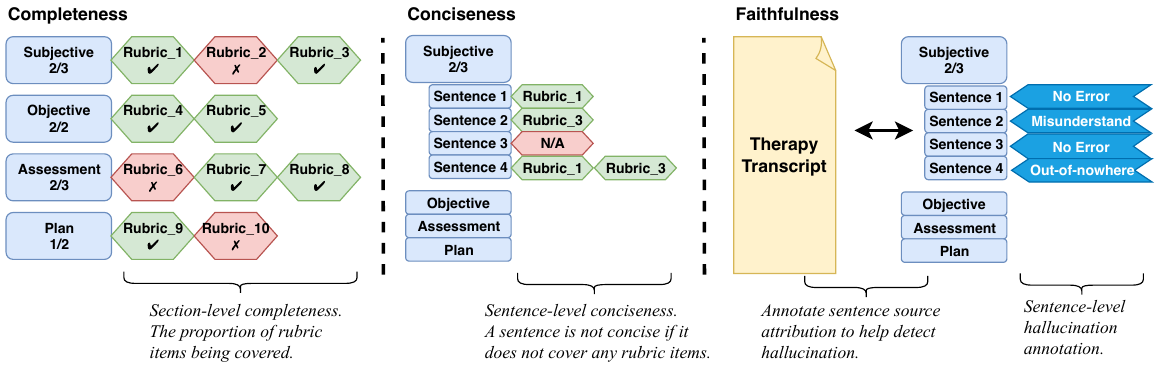}
    \caption{The \humaneval human evaluation protocol.}\vspace{-2ex} 
    \label{fig:protocol}
\end{figure*}

\section{SOAP Note and Rubric creation}\label{sec:rubric}

In \papername, we look at a popularly used therapy note documentation format: SOAP, an acronym for Subjective, Objective, Assessment, Plan, with each letter representing a section of the note \cite{weed1964medical}. 
At a high level, in SOAP notes, the \textit{subjective} component consists of insights about the client's presenting problem from the client's viewpoint and that of significant others. In contrast, the \textit{objective} component includes the counselor's observations. The \textit{assessment} section shows how the subjective and objective data are being analyzed, interpreted, and considered, and the \textit{plan} section outlines the treatment approach~\cite{cameron2002learning}. While there exist other therapy note formats, we use SOAP notes because they are widely referenced in behavioral therapy \cite{berghuis2014adolescent, reiter2023writing}, standardized in major electronic health records \cite{podder2022soap,gao2023progress}, and provide a representative framework for developing better evaluation protocols.

In practice, the exact definitions and information present in each of the sections are determined by the healthcare provider organization and their record management practices. Therefore, a fine-grained set of consistent rubrics is necessary to complement the definition. In addition to the generally underspecified definitions, there is a lack of consistent clinical psychology literature for best practices in writing therapy notes and key characteristics that determine the quality of a note. To determine what \textit{high-quality means to domain experts}, we work with five therapists to \textbf{co-design a rubric consisting of the different section-wise dimensions of note quality.} 


\subsection{Domain Experts}
To develop the rubric, we collaborated with \textbf{Therapist A}\footnote{Therapist A is a co-author of this paper} who has over 20 years of clinical experience. Additionally, we worked with four other therapists from diverse professional backgrounds who hold a Psy. D., Ph.D. in counseling psychology, or licensed clinical social work and have experience with multiple healthcare providers, as well as training new therapists in therapeutic techniques and note-writing. 

\subsection{Rubric Creation Procedure}
Our co-design consists of a two-step process. In the first step, we conduct three hour-long sessions with Therapist A to identify key characteristics of each SOAP note section, assign their relative importance, and refine the rubric through iterative feedback and example notes, including general section-agnostic guidelines. 

In the second step, we ask four more therapists to verify the section appropriateness and the relative importance of each key characteristic. We also ask the therapists to suggest key characteristics that may be missed from the first step. The process is completed in an annotation tool shown in appendix figure~\ref{fig:rubric_ann_tool}. After the annotation, we consolidate the rubric by taking the majority vote. The final definitions of the SOAP note and the corresponding section-wise key characteristics are presented in appendix~\ref{soap_note_rubric}. We also validate the final rubric with ($N=17$) external therapists employed in note writing ($N=8$) and evaluation ($N=9$), as mentioned in appendix \ref{sec:annotator_qualification}. 

\textbf{Rubric Quality:} 
We observe perfect \ac{IAA} among 5 experts for the appropriate section for each key characteristic and observe high agreements for the relative importance of each key characteristic -- Fleiss' $\kappa$: 0.68, Krippendorff's $\alpha$: 0.73. Detailed \acp{IAA} by section are shown in appendix Table \ref{table:kappa_alpha_comparison}. 

\section{Evaluation Protocols}\label{sec:eval}
In this section, we introduce human and automatic evaluation protocols using the rubric, denoted as \humaneval and \autoeval, respectively. Both focus on three dimensions: 
%

\noindent\textbf{Completeness:} This dimension evaluates whether each rubric element appears in its corresponding note section (e.g., the chief complaint in subjective). The score is computed as the ratio of covered elements aggregated across sections.

\noindent\textbf{Conciseness:} This metric measures whether each sentence contributes to a rubric item. Annotators (human or automated) label sentences accordingly, and the score is the ratio of necessary sentences in a section.

\noindent\textbf{Faithfulness:} This evaluation checks whether a note's content is factually grounded in the therapy session. Errors are categorized into hallucination types, ensuring a granular assessment.

\noindent\textbf{Why these three dimensions?} 
The evaluation dimensions were chosen based on practical considerations and therapist feedback. Since no standardized framework exists for grading therapists' notes, completeness is crucial to meet regulatory requirements. Therapists also emphasized conciseness, noting concerns about AI-generated verbosity. Lastly, faithfulness was included to mitigate hallucinations in LLM-generated text, ensuring accuracy and reliability.

\subsection{Human Evaluation Protocols} \label{sec:method_human}
Our \humaneval relies on our rubric design to break down each dimension into more objective, simpler, and cost-effective tasks. Figure~\ref{fig:protocol} illustrates the human evaluation protocol.

To find the \textbf{completeness}, a therapist reviews a note section and marks covered rubric elements. The full note score is computed as a micro average, weighted by section rubric elements. This design minimizes annotators’ effort in reviewing lengthy therapy transcripts (~45 minutes).
For \textbf{conciseness}, annotators label sentences with relevant rubric items or mark them as unnecessary. This annotation is done separately from completeness to prevent biased coverage assessment.
In the case of \textbf{faithfulness}, annotators cross-check sentences against the therapy transcript, selecting supporting content from the source and categorizing hallucinations into (1) Out-of-nowhere, (2) Misinterpreted Information, or (3) No Error. Given the session length, this is the most costly evaluation.
\textit{While non-experts could perform this task, all annotations in our study are conducted by licensed U.S. therapists to ensure accuracy and reliability.}

\subsection{Automatic Evaluation Protocols} \label{sec:method_auto}
We use LLMs to mimic human annotators to get the completeness and conciseness evaluation. For \textbf{completeness}, we present a note and one rubric item to an LLM and ask if the item appears in the summary. For \textbf{conciseness}, we break down the note into sentences, and for each sentence, we verify if a rubric element is covered in the sentence. 
We use AlignScore~\citep{zha-etal-2023-alignscore} for the \textbf{faithfulness} evaluation.

\begin{table*}[tb]
\centering\small
\begin{tabular}{lccccccc}
\toprule
                 & \multicolumn{2}{c}{\textbf{Completeness}} & \multicolumn{2}{c}{\textbf{Conciseness}} & \multicolumn{2}{c}{\textbf{Faithfulness}} & \textbf{Acceptance} \\\cmidrule(lr){2-3} \cmidrule(lr){4-5} \cmidrule(lr){6-7} \cmidrule(lr){8-8}
\textbf{Note}             & \textbf{\humaneval}          & \textbf{Likert}         & \textbf{\humaneval}         & \textbf{Likert}         & \textbf{\humaneval}          & \textbf{Likert}         & \textbf{Likert}       \\\midrule
Human            & 29.5 (±12.4)    & 2.85 (±1.09)   & 75.6 (±14.9)   & 4.28 (±0.89)   & 87.0 (±12.6)    & 4.43 (±0.81)   & 2.34 (±0.75) \\
Llama 3.1 70B    & 39.7 (±7.9)     & 3.80 (±0.79)   & 84.0 (±12.1)    & 4.83 (±0.35)   & 68.5 (±15.1)    & 4.68 (±0.50)   & 3.34 (±0.61) \\
Mistral Large V2 & 38.1 (±7.5)     & 4.01 (±0.70)   &  91.5 (±7.1)   & 4.88 (±0.35)   & 71.8 (±14.0)    & 4.90 (±0.34)   & 3.73 (±0.70) \\
\bottomrule
\end{tabular}
\caption{Human evaluation results using \humaneval and Likert human evaluations. The values in brackets show the standard deviation over the 50 examples. ``Acceptance'' refers to the overall acceptance annotated on a Likert scale. This table shows aggregated scores for the full note. See Appendix Table~\ref{tab:human_eval_section} for a breakdown by sections.}\label{tab:human_eval}\vspace{-2ex}
\end{table*}

\section{Data collection and note generation}

\noindent\textbf{Dataset:}
We conducted experiments on therapy conversations from the AnnoMI dataset~\cite{annomi2}. Due to the cost of recruiting expert therapists for annotation, we chose the first 50 conversations from the high-quality split of AnnoMI (the median conv. len. = 1067 words/ 42 turns). 

\noindent\textbf{Human Note Collection:} The notes were written by the $N=5$ internal therapists involved in the rubric design, and we also recruited ($N=8$) therapists to write notes for these 50 conversations. The cost to collect each note was \$206.

\noindent\textbf{LLM Note Generation:}
We prompted several off-the-shelf LLMs to generate notes, including Claude~\cite{anthropic2024claude}, Llama~\cite{touvron2023llama} and Mistral~\cite{jiang2023mistral}, and also use two clinical and therapy domain adapted LLMs -- MentalLlama\cite{yang2024mentallama} and OpenBioLLM \cite{OpenBioLLMs}. Appendix~\ref{sec:prompt_note} shows the prompt we used for note generation. The prompt is simple and not carefully optimized for any particular LLM to achieve a fair comparison between LLMs.

\noindent\textbf{Human Evaluation:} For evaluation, we recruited $N=9$ external therapists who are different from those who wrote the notes. The cost to collect a single human evaluation related to one note is $\$190$. We followed the \humaneval protocol described in Section~\ref{sec:method_human}, and collected two independent annotations for each note. We also collect annotations for \textit{5-point Likert-scale baseline} on three aspects -- Completeness, Conciseness, and Faithfulness. Experts also annotate the \textit{overall acceptance} of a note on a scale of 1 to 5. Due to the high cost of human annotation, we only conducted the \humaneval on human notes and 2 LLM generated notes -- Llama 3.1 (70B) and Mistral Large V2.

\noindent\textbf{Automatic Evaluation:} We followed the protocol in Section~\ref{sec:method_auto} to conduct automatic evaluation. We also explored a Likert-style automatic evaluation similar to LLM-as-a-judge~\cite{zheng2023judging} (refer to appendix~\ref{sec:auto_eval_prompts} for corresponding prompts). We compared \papername with conventional reference-based evaluation, such as ROUGE~\cite{lin-2004-rouge} and BERTScore~\cite{zhang2019bertscore}, and we find the efficacy of our automatic evaluation protocol by correlating the automatic metric with human annotations at the note-level.

\section{Experiments}
\subsection*{Q1: How reliable is \humaneval compared to the conventional Likert-based approach?}\label{sec:q1}
Table~\ref{tab:human_iaa} shows the \ac{IAA} between two annotators for each type of human rating we collect. We found that Krippendorff's $\alpha$ for \humaneval is significantly higher than that of the Likert-style evaluation for all three dimensions, showing that \humaneval can achieve more consistent annotations across two independent annotators and is thus more reliable. Furthermore, \humaneval provides distinct variance in outputted scores as compared to expert Likert scale judgments (refer figures \ref{human_tn_var}, \ref{human_likert_var}). 

\begin{table}[htb]
\small\centering
\begin{tabular}{lcccc}
\toprule
                      & \multicolumn{2}{c}{\textbf{\humaneval}}     & \multicolumn{2}{c}{\textbf{Likert}} \\\cmidrule(lr){2-3}\cmidrule(lr){4-5}
\textbf{Dimension}      & \textbf{Raw Agg.} & \textbf{K-$\alpha$} & \textbf{MSE}  & \textbf{K-$\alpha$} \\\midrule
\textbf{Completeness}  & 77.6         & 0.52       & 2.72       & 0.08          \\
\textbf{Conciseness}  & 85.5         & 0.49       & 1.01       & 0.16          \\
\textbf{Faithfulness} & 85.9         & 0.62       & 0.86       & 0.18          \\
\textbf{Acceptance}   & -            & -          & 2.24       & 0.15          \\\bottomrule
\end{tabular}
\caption{\ac{IAA} of human evaluations. We show raw agreement and Krippendorff's $\alpha$ (K-$\alpha$) for rubric annotations and mean squared error (MSE) and K-$\alpha$ for Likert annotations. ``Acceptance'' refers to the overall acceptance annotated on a Likert scale. \humaneval appears to have better annotation consistency compared to Likert annotations.}\label{tab:human_iaa}
\end{table}
\begin{table*}[tb]
\small\centering
\begin{tabular}{c|lcccccc}
\toprule
\multicolumn{2}{c}{} & \multicolumn{2}{c}{\textbf{Completeness}} & \multicolumn{2}{c}{\textbf{Conciseness}} & \multicolumn{2}{c}{\textbf{Faithfulness}} \\\cmidrule(lr){3-4}\cmidrule(lr){5-6}\cmidrule(lr){7-8}
\textbf{Evaluator} & \textbf{Note Source}                & \textbf{\autoeval}           & \textbf{Likert}         & \textbf{\autoeval}          & \textbf{Likert}         & \textbf{\autoeval}           & \textbf{Likert}         \\\midrule
Claude 3  & Human                      & 26.4 (±12.4)    & 2.85 (±0.41)   & 63.4 (±22.2)   & 3.15 (±0.49)   & 73.2 (±14.9)    & 4.11 (±0.57)   \\
Sonnet    & Claude 3 Sonnet            & 34.8 (±7.3)     & 3.39 (±0.27)   & 86.0 (±10.3)   & \textbf{3.86 (±0.21)}   & 74.0 (±10.1)    & 4.73 (±0.36)   \\
          & Claude 3 Haiku             & 36.8 (±8.7)     & 3.46 (±0.27)   & \textbf{87.6 (±12.1)}   & 3.81 (±0.21)   & 69.9 (±10.1)    & 4.70 (±0.38)   \\
\scriptsize{(AlignScore}  & Llama 3.1 (70B)            & 35.1 (±8.0)     & 3.33 (±0.36)   & 84.8 (±12.5)   & 3.52 (±0.27)   & 69.0 (±11.6)    & 4.49 (±0.50)   \\
\scriptsize{for} & Llama 3.1 (8B)             & 35.0 (±6.8)     & 3.22 (±0.27)   & 85.9 (±8.6)    & 3.55 (±0.27)   & 70.2 (±11.5)    & 4.63 (±0.44)   \\
\scriptsize{Faithfulness)} & Mistral Large V2           & 36.8 (±8.2)     & 3.50 (±0.32)   & 84.3 (±9.1)    & 3.83 (±0.20)   & 75.8 (±8.8)     &  4.91 (±0.20)   \\
          & Mistral (7B)               & \textbf{37.7 (±8.6)}     & \textbf{3.58 (±0.28)}   & 81.2 (±10.5)   & 3.85 (±0.17)   & 75.2 (±9.5)     & \textbf{4.93 (±0.19)}   \\
          & MentaLlama (13B)               & 24.5 (±10.2)     & 2.86 (±0.33)  & 77.0 (±20.8)   & 3.42 (±0.40)    & \textbf{80.4 (±9.9)}     & 4.50 (±0.60)   \\
          & OpenBioLLM (70B)               & 24.6 (±9.9)     &  3.19 (±0.42) & 72.9 (±13.9)   &  3.72 (±0.45)  & 80.0 (±11.0)     & 4.76 (±0.55)   \\
\bottomrule
\end{tabular}
\caption{\autoeval and Likert-style automatic evaluation. We show the results using Claude 3 Sonnet as the evaluator. Note that the \autoeval faithfulness is not an LLM-based metric, instead, it uses AlignScore. }\label{tab:reference_free}
\vspace{-2ex}
\end{table*}

Table~\ref{tab:human_eval} shows human annotated scores for human written notes and 2 LLM generated notes. Note that the sources are revealed to the annotators. \textbf{It is surprising to see, according to Likert-style scores, that experts judge LLM-generated notes to be superior to human written notes across all dimensions -- completeness, conciseness, faithfulness, and overall acceptance.}
Our \humaneval shows the same order for completeness and conciseness, however, for faithfulness, \humaneval shows higher scores for human-generated notes, which shows the advantage of using our rubric,  breaking down each section into smaller, more objective annotation tasks.  


\subsection*{Q2: Does \autoeval align with human and reference-based automatic evaluations?}\label{sec:q2}


We find that all traditional reference-based metrics show similar values, making these n-gram-based metrics insufficient to provide meaningful signals for generation quality (refer appendix table ~\ref{tab:reference_based}). Next, table~\ref{tab:hm_corr} shows the correlation between two sets of automatic evaluation (\autoeval and Likert-style LLM-as-a-Judge) and two sets of human evaluation (\humaneval and Likert-scores). \textbf{Notably, \autoeval and \humaneval indicated the highest correlation, as shown in Column (A), demonstrating that the fine-grained evaluation achieves higher agreement between human and LLM evaluators.}
Column (B) reveals the utility of the LLM-as-a-Judge for the completeness evaluation but presents a poor correlation for conciseness and faithfulness. When comparing automatic evaluations with Human Likert-scale annotations, the correlations appear to be generally poor, suggesting that neither of the automatic evaluations correlates well with human Likert-scale evaluation. Overall, the faithfulness correlation shows significant challenges in hallucination detection. 
Human and automatic evaluations agree that human written notes are roughly 10\% less complete and 10\% less concise compared to LLM-generated notes. For the faithfulness evaluation, humans appear to favor human written notes, while automatic evaluations favor LLM notes.

\begingroup
\setlength{\tabcolsep}{2pt}
\begin{table}[tb]
\small\centering
\begin{tabular}{c|rcc|cc}
\toprule

 & \textbf{v.s.}              & \multicolumn{2}{c}{\textbf{\humaneval}} & \multicolumn{2}{c}{\textbf{Human Likert}} \\
 & \scriptsize \textbf{Evaluator}         & \scriptsize \textsuperscript{(A)}\textbf{\autoeval}   & \scriptsize \textsuperscript{(B)}\textbf{Likert}     & \scriptsize \textsuperscript{(C)}\textbf{\autoeval}   & \scriptsize \textsuperscript{(D)}\textbf{Likert}    \\\midrule
 \multirow{3}{*}{\rotatebox[origin=c]{90}{\textbf{Comp.}}} 
 & Claude 3 Sonnet  & \textbf{0.58}            & 0.46      & 0.24                      & 0.34      \\
 & Llama 3.1 (70B)  & 0.44                     & 0.55      & 0.23                      & 0.36      \\
 & Mistral Large V2 & 0.48                     & 0.55      & 0.34                      & 0.36      \\\midrule
\multirow{3}{*}{\rotatebox[origin=c]{90}{\textbf{Conc.}}} 
 & Claude 3 Sonnet  & 0.36                     & 0.27      & 0.19                      & 0.26      \\
 & Llama 3.1 (70B)  & 0.39                     & 0.14     & 0.26                      & 0.11      \\
 & Mistral Large V2 & \textbf{0.40}                     & 0.24      & 0.21                      & 0.17      \\\midrule
\multirow{3}{*}{\rotatebox[origin=c]{90}{\textbf{Faith.}}} 
 & Claude 3 Sonnet  &  -                       & -0.15     & -                        & 0.28      \\
 & Llama 3.1 (70B)  &  -                       & -0.20     & -                         & 0.18      \\
 & Mistral Large V2 &  -                       & -0.22     & -                         & 0.19      \\
 & AlignScore       &  \textbf{0.34}           & -         & 0.27                     & -      \\
\bottomrule
\end{tabular}
\caption{The note-level correlation between automatic metrics and human annotations. Column (A) and (B) compares automatic evaluation with \humaneval. \autoeval achieves much higher correlation than Likert-style LLM-as-a-Judge. Column (C) and (D) compares automatic evaluation with human Likert-style annotation, where the correlation is generally poor. }\label{tab:hm_corr}
\vspace{-3ex}
\end{table}
\endgroup

\subsection*{Q3: How effectively do LLMs generate notes?}\label{sec:q3}


\emph{Upon asking experts to rate notes without telling the source of the note (refer table \ref{tab:human_eval}), we observe that experts prefer and judge LLM-generated notes to be superior to human-written notes across all dimensions except fine-grained faithfulness evaluation.} This highlights the potential of using LLMs for therapy note construction.

\noindent \textbf{Note Length: }For further investigation, we examine the length of notes written by the therapists and LLMs (appendix table~\ref{tab:note_length}). Human written notes are generally shorter, and in particular, the ``plan'' section of human notes is much shorter than LLM notes (Average length of human-plan section notes = 29 words, Average length of LLM-generated plan section = 78.5 words). This is because therapists tend to be very concise, with just one sentence stating the follow-up session, while LLM-generated notes contain more content such as ``short-term goals'' and ``long-term goals'' (see table~\ref{tab:concise_example}). We believe that the natural and fluent English writing from LLMs likely biases human annotators, thus conflating fluency with accuracy~\cite{elangovan-etal-2024-considers}. 
Next, we manually observed some examples (appendix table~\ref{tab:concise_example}) and found that humans tend to write shorter sentences for the same rubric items. Based on a subsequent conversation with Therapist A, we uncover that therapists spend substantial time with various kinds of documentation and find themselves hard-pressed to write descriptive quality notes \cite{griswold2019notes}. 

\noindent \textbf{Section-wise scores: }On analyzing
the  \humaneval score breakdown for each rubric item, we observe that human written notes show considerably less coverage for some rubric items (refer to appendix table ~\ref{rab:rubric_detail}). For example, ``symptoms'' in the Subjective, ``mental status'' in the Objective, and ``future interventions'' in the Plan show a large discrepancy ( more than $~20 \%$). 

\noindent \textbf{Automatic evaluation: }We show the automatic evaluation results on Human notes and several LLM notes in Table~\ref{tab:reference_free}. The numbers reflect a similar pattern as the human evaluation, where LLMs, in general, outperform humans in note completeness and conciseness. Among LLMs, Mistral tends to be more conservative, with more faithful content.

\section{Conclusion}
In this paper, we conducted analyses on quality evaluation strategies for behavior health therapy notes. By collaborating with domain experts to design a rubric, we designed fine-grained human evaluation and automatic evaluation protocols. We demonstrated the advantage of \humaneval against conventional Likert-style human evaluation. Expert evaluation with \humaneval and conventional Likert-scales shows preference towards LLM-generated notes. Our \autoeval outperformed the conventional LLM-as-a-Judge strategy and showed a higher correlation with human evaluations for completeness and conciseness, while the faithfulness evaluation remains a challenge. Thus, we urge research toward robust and automatic evaluation of therapy notes. Subsequently, we are sharing high-quality note annotations from practitioners, the co-designed rubric, and all annotations we collected in the project to benefit the research community.

\section*{Ethical Considerations}
The organization's review protocols approved the current study. We do not advocate for fully automated LLM-generated notes; rather, we propose augmenting therapist workflows by providing an LLM-generated draft as a starting point. Furthermore, all therapy transcripts used in this work are from an open-source dataset -- AnnoMI ~\cite{annomi2}.  Lastly, to ensure the appropriate stake-holder inclusion and the generalizability of findings, ($N=22$) therapists were consulted in this study in the following capacities:
\begin{enumerate}
    \item Therapist involvement in the co-design process. We work with $N=5$ senior therapists from one of the largest behavioral healthcare networks in the country.
    \item Therapist involvement in note construction (annotation). For this step, we engage with $N=5$ of the therapists mentioned above and additionally with $N=8$ therapists from another organization.
    \item Therapist involvement in note evaluation. For this step, we worked with $N=9$ new therapists, which were separate from the previous two groups. These nine therapists evaluated SOAP notes with the help of the rubric.
\end{enumerate}

\textbf{Deployment:} Automated behavioral health note generation and evaluation tools in real-world settings necessitate compliance with HIPAA (Health Insurance Portability and Accountability Act) regulations and privacy-preserving AI practices. Thus, we include open-source models for both -- note generation and evaluation, so as to show results for models which can be run on private compliant servers.
\bibliography{main_new}

\clearpage
\appendix
\onecolumn
\section*{Appendix}

\section{Definitions of SOAP note sections}
\label{soap_note_rubric}
\subsection{Subjective}
\textbf{Definition:} In this section, document the subjective reports from the client, their family members, and past medical records. Include how the client describes their feelings and current symptoms.

\paragraph{Key Characteristics:} 
\begin{itemize}[leftmargin=*]
    \setlength\itemsep{0em}
    \item \textbf{Chief Complaint:} The reason why the client is seeking therapy. Could also be a description of what symptoms the client is experiencing. \textbf{Importance: }\textcolor{blue}{ Mandatory}
    \item \textbf{Symptoms (as the client is talking about it):} The client’s own description of their feelings, thoughts, and behaviors along with the severity. \textbf{Importance: }\textcolor{blue}{ Mandatory}
    \item \textbf{History:} Relevant background information, including any past medical, therapy, or behavioral issues. \textbf{Importance: }\textcolor{blue}{ Mandatory}
    \item \textbf{Client's Goals:} What the client hopes to achieve through therapy. \textbf{Importance: }\textcolor{blue}{ Highly recommended}
    \item \textbf{Homework from Previous Sessions:} Reviewing homework from the previous sessions and noting the client’s compliance. \textbf{Importance: }\textcolor{blue}{ Highly recommended}
    \item \textbf{Quotes:} Direct quotes from the client can be particularly useful to capture their exact words and emotional tone. \textbf{Importance: }\textcolor{blue}{ Highly recommended}
\end{itemize}

\subsection{Objective}
\textbf{Definition:} This section is for recording objective observations made during the session. Note any factual, observable information, such as the client's appearance, behavior, mood, affect, and speech patterns. Avoid including any subjective statements or self-reported information from the client.

\paragraph{Key Characteristics:} 
\begin{itemize}[leftmargin=*]
    \setlength\itemsep{0em}
    \item \textbf{Client’s Observed Behavior:} The therapist's observations of the client's behavior, mood, appearance, and affect during the session. \textbf{Importance: }\textcolor{blue}{ Mandatory}
    \item \textbf{Mental Status:} Observations regarding the client’s appearance, speech, thought processes, and orientation. \textbf{Importance: }\textcolor{blue}{ Mandatory}
    \item \textbf{Assessment Tools:} Results from any standardized assessments or scales used during the session. \textbf{Importance: }\textcolor{blue}{ Highly recommended}
    \item \textbf{Therapy Activities:} Description of specific interventions or activities conducted during the session. \textbf{Importance: }\textcolor{blue}{ Highly recommended}
    \item \textbf{Interventions [A]:} Applied interventions and treatment plans (MI, Cognitive Restructuring, DBT, etc.). \textbf{Importance: }\textcolor{blue}{ Highly recommended}
    \item \textbf{Interventions [B]:} Focus on describing active interventions provided rather than passive ones. \textbf{Importance: }\textcolor{blue}{ Highly recommended}
\end{itemize}

\subsection{Assessment}
\textbf{Definition:} In this section, integrate the subjective and objective information to provide a comprehensive analysis of the client's current condition. Summarize the clinical impressions and hypotheses regarding the client's issues.

\paragraph{Key Characteristics:} 
\begin{itemize}[leftmargin=*]
    \setlength\itemsep{0em}
    \item \textbf{Diagnosis/Symptoms:} Any formal diagnoses made based on DSM-5 criteria or other diagnostic tools. \textbf{Importance: }\textcolor{blue}{ Mandatory}
    \item \textbf{Identifying Triggers:} Any triggers shown by the client.
    \item \textbf{Progress:} Evaluation of the client's progress toward their therapeutic goals. \textbf{Importance: }\textcolor{blue}{ Highly recommended}
    \item \textbf{Analysis:} The therapist's interpretation of how the client's subjective report and objective observations relate to their overall condition. \textbf{Importance: }\textcolor{blue}{ Highly recommended}
    \item \textbf{Response to Interventions.} \textbf{Importance: }\textcolor{blue}{ Highly recommended}
    \item \textbf{Overall/High-Level Progress.} \textbf{Importance: }\textcolor{blue}{ Highly recommended} 
    \item \textbf{Treatment Goals:} Specific, measurable, achievable, relevant, and time-bound (SMART) goals for the client. Adjustments to the treatment goals. \textbf{Importance: }\textcolor{blue}{ Highly recommended}
    \item \textbf{Stages of Change:} For interventions like Motivational Interviewing, note the client's stage of change (Pre-contemplation, Contemplation, Action, Maintenance, etc.). \textbf{Importance: }\textcolor{blue}{ Highly recommended}
\end{itemize}

\subsection{Plan}
\textbf{Definition:} Outline the next steps for the client's treatment. Include both short-term and long-term goals, specifying what will be addressed in the next session as well as overall treatment objectives. 

\paragraph{Key Characteristics:} 
\begin{itemize}[leftmargin=*]
    \setlength\itemsep{0em}
    \item \textbf{Future Interventions:} Planned therapeutic techniques or strategies to be used in future sessions. \textbf{Importance: }\textcolor{blue}{ Mandatory}
    \item \textbf{Follow-Up:} Scheduling of the next session and any referrals to other professionals if needed. Note the date for the next appointment if decided upon. \textbf{Importance: }\textcolor{blue}{ Mandatory}
    \item \textbf{Adjustment of Medication/Intervention Choice.} \textbf{Importance: }\textcolor{blue}{ Mandatory in certain circumstances} 
    \item \textbf{Homework:} Assignments or activities for the client to work on between sessions. \textbf{Importance: }\textcolor{blue}{ Highly recommended}
\end{itemize}

\subsection{General Items}
\paragraph{Key Characteristics:} 
\begin{itemize}[leftmargin=*]
    \setlength\itemsep{0em}
    \item Clearly reflect that the practitioner assessed for and addressed any safety concerns (e.g., suicide risks, self-harming behaviors, homicidal ideation, etc.). \textbf{Importance: }\textcolor{blue}{ Mandatory}
    \item Evidence of treatment being provided in a culturally competent manner. \textbf{Importance: }\textcolor{blue}{ Highly recommended}
    \item \textbf{Professionalism} \textbf{Importance: }\textcolor{blue}{ Highly recommended}
    \begin{itemize}
        \item Never describe other clients and staff in a derogatory manner.
        \item Avoid using slang.
        \item Descriptions of the patient’s presenting problem should be used rather than presumptuous adjectives.
    \end{itemize}
\end{itemize}

\section{Limitations}

\noindent\textbf{Real data availability :} Because of the sensitive nature of behavioral health, real doctor-patient conversations are confidential. The public data we use in this study appears to be shorter and less complex than a real therapy session. 

\noindent\textbf{The scale of study:} This study is relatively small due to the cost of recruiting licensed therapists, involving only two open-weight LLMs and human-written notes across a limited number of conversations. Annotator bias may also affect results. While differences between human and LLM notes are clear, the gap between the two LLMs is small, and their ranking may vary across different datasets.

\noindent\textbf{LLM performance:} We used simple prompts in this study, focused on evaluating the framework rather than optimizing LLM performance. The results could likely improve with more advanced prompt engineering.


\section{Annotator Qualification and Cost}
\label{sec:annotator_qualification}
\textbf{Human Note Writing}: Human notes were written by the $N=5$ internal therapists involved in the rubric design, as well as $N=8$ external therapists. All external therapists hold either a Master's or Ph.D. degree in clinical psychology, and are licensed therapists or clinical social workers in the United States, with experience ranging from 3 to 18 years. The cost to collect each note was \$206.

\noindent\textbf{Note Evaluation}: Human evaluation was conducted by $N=9$ external therapists who are different from those who wrote the notes. All evaluators are licensed therapists or clinical social workers with a Master's or Ph.D. degree in a related field. The cost to collect a single human evaluation related to one note is $\$190$.

\noindent\textbf{Total cost:} Annotating a large number of conversations with highly specialized experts is time-consuming and costly. The cost of collecting one note for each conversation was \$206, making the cost of the dataset creation to be \$10,300. We incur additional costs in the human evaluations (\$190 for each, 150 evaluations total). This makes our total cost to be \$38800, limiting the size of the dataset to 50 conversations.

\section{Human Rubric Creation Details}
\begin{figure*}[htb]
    \centering
    \includegraphics[width=\textwidth]{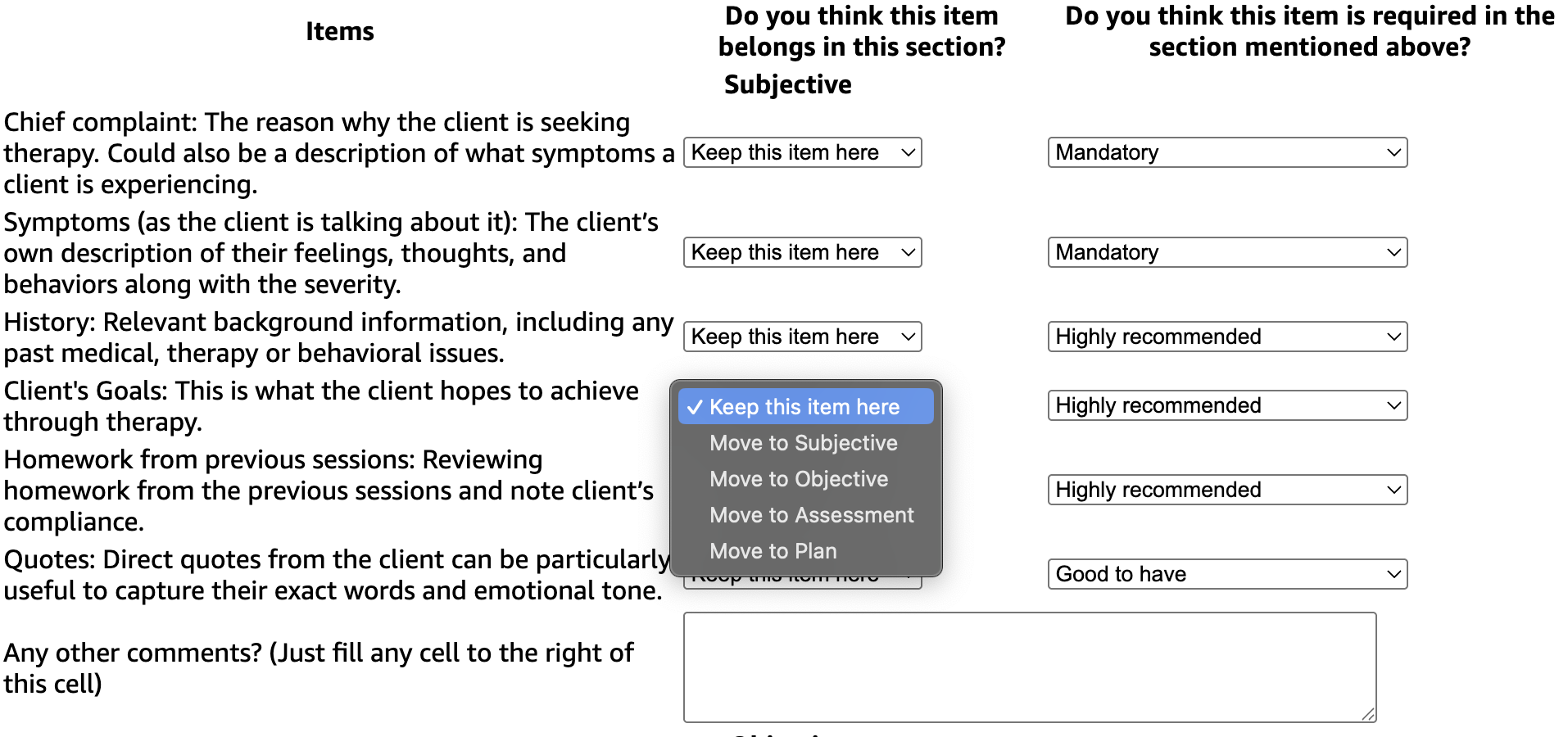}
    \caption{Rubric annotation tool. For each rubric, a therapist would read it and annotate (1) if the section is appropriate and (2) the importance level.}
    \label{fig:rubric_ann_tool}
\end{figure*}

Figure~\ref{fig:rubric_ann_tool} shows the interface of the tool used to build the rubric. 

\section{Automatic Evaluation Details}
\subsection{Prompts for \autoeval} \label{sec:auto_eval_prompts}
\subsubsection*{Rubric-based Completeness Evaluation}
\begin{lstlisting}
Below is a behavioral therapy progress note segment. The rubric item outlines one of the necessary components for the note. Verify if the rubric item presents in the progress note segment. 

## Note Segment
{note_segment}

## Rubric Item (an item that should present in the note segment)
{rubric_item}

Does the note segment contain the rubric item? Response in [Yes, No] with no other content:
\end{lstlisting}

\subsubsection*{Rubric-based Conciseness Evaluation}
\begin{lstlisting}
Below is a sentence from a behavioral therapy progress note. The rubrics outlines the necessary components for the note. Verify if the note sentence fit in one of the rubric items.

## Note Sentence
{note_sentence}

## Rubrics (a list of items that should present in the note segment)
{rubrics}

Does the note sentence fit in one of the rubric items? Response in [Yes, No] with no other content:
\end{lstlisting}

\subsection{Prompts for Likert-style automatic evaluation}
\subsubsection*{Completeness}
\begin{lstlisting}
Below is a behavioral therapy conversation along with a corresponding progress note segment. The rubrics outline the necessary components for the note. Based on the conversation and rubrics, evaluate the completeness of the note segment.

## Conversation
{conversation}

## Note Segment
{note_segment}

## Rubrics (a list of items that should present in the note segment)
{rubrics}

## Rating Codebook
1: The note segment is missing most of the key information from the conversation.
2: The note segment includes some important details but is significantly incomplete.
3: The note segment contains a moderate amount of important information.
4: The note segment captures most of the key information from the conversation.
5: The note segment comprehensively captures all the key information.

Using the 1 to 5 scale from the rating codebook, rate the completeness of the note segment. Output only the rating [1, 2, 3, 4, 5]:
\end{lstlisting}

\subsubsection*{Conciseness}
\begin{lstlisting}
Below is a behavioral therapy conversation along with a corresponding progress note segment. The rubrics outline the necessary components for the note. Based on the conversation and rubrics, evaluate the conciseness of the note segment.

## Conversation
{conversation}

## Note Segment
{note_segment}

## Rubrics (a list of items that should present in the note segment)
{rubrics}

## Rating Codebook
1: The note segment includes substantial non-important information that detracts from the main points.
2: The note segment includes non-important information that needs to be reduced.
3: The note segment includes some non-important information but does not heavily detract from the main points.
4: The note segment includes minor non-critical information.
5: The note segment includes no non-important information, making it concise and focused.

In the scale of 1 to 5, rate the conciseness of the note segment following the rating codebook. Output only the rating [1, 2, 3, 4, 5]:
\end{lstlisting}

\subsubsection*{Faithfulness}
\begin{lstlisting}
Below is a behavioral therapy conversation along with a corresponding progress note segment. Verify the faithfulness of the note segment based on the conversation.

## Conversation
{conversation}

## Note Segment
{note_segment}

## Rating Codebook
1: The note segment contains significant inaccuracies or false information.
2: The note segment contains several inaccuracies or false information.
3: The note segment may contain some inaccuracies or false information.
4: The note segment contains minor non-critical inaccuracies or false information.
5: The note segment contains no inaccuracies or false information.

In the scale of 1 to 5, rate the faithfulness of the note segment following the rating codebook. Output only the rating [1, 2, 3, 4, 5]:
\end{lstlisting}

\section{Prompt for Note Generation}\label{sec:prompt_note}
\begin{lstlisting}
In emotional support conversations, two primary roles exist: the therapist (individual providing support) and the client (individual seeking support). Your task is to summarize an emotional support conversation into client progress notes. These notes are usually in the SOAP format. The SOAP is a standardized form of recording a client's progress. It stands for:

- Subjective: In this section, document the subjective reports from the client, their family members, and past medical records. Include how the client describes their feelings and current symptoms.
- Objective: This section is for recording objective observations made during the session. Note any factual, observable information, such as the client's appearance, behavior, mood, affect, and speech patterns. Avoid including any subjective statements or self-reported information from the client. 
- Assessment: In this section, integrate the subjective and objective information to provide a comprehensive analysis of the client's current condition. Summarize your clinical impressions and hypotheses regarding the client's issues. 
- Plan: Outline the next steps for the client's treatment. Include both short-term and long-term goals, specifying what will be addressed in the next session as well as overall treatment objectives. Be clear and specific about your expectations and the client's goals for the duration of treatment.

Output Dictionary template: 
{
"Subjective": "...",
"Objective": "...",
"Assessment": "...",
"Plan": "..."
}
Generate notes for the provided conversation in the above Dictionary style template. 

{Conversation}

SOAP Note: 
\end{lstlisting}
\clearpage

\section{Human label distribution}

\begin{figure}[htb]
\centering
\includegraphics[width=\textwidth]{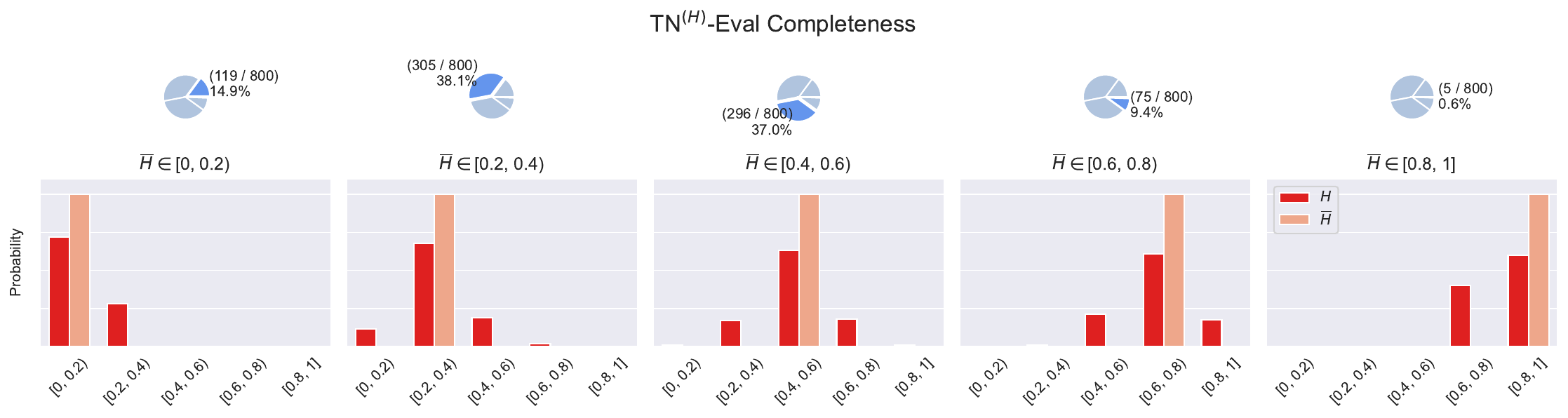}\\
\noindent\rule{0.8\textwidth}{1pt}
\includegraphics[width=\textwidth]{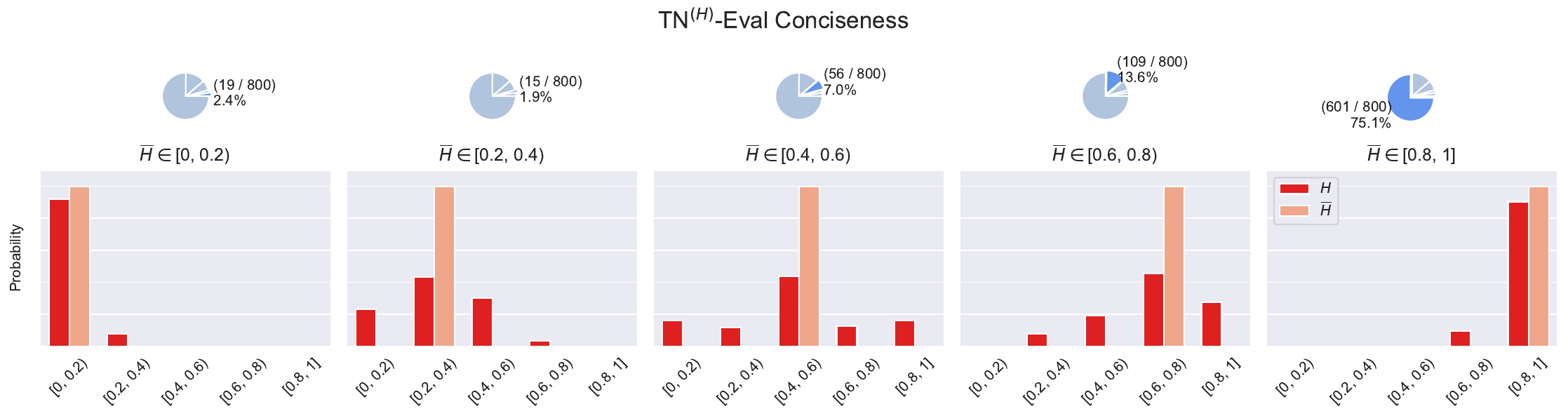}\\
\noindent\rule{0.8\textwidth}{1pt}
\includegraphics[width=\textwidth]{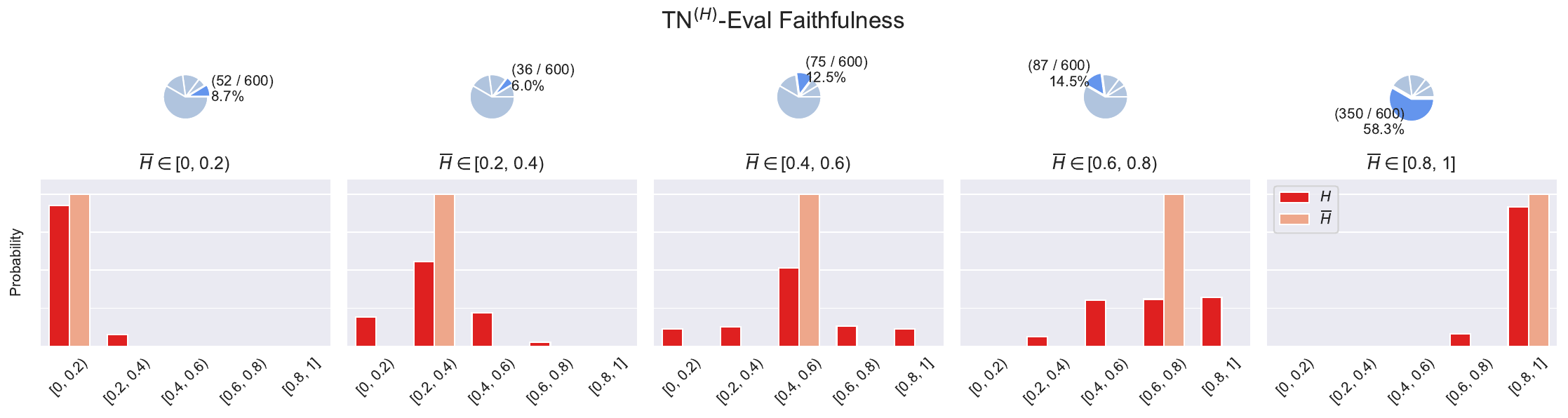}
\caption{Human label distribution for \humaneval  annotations.}
\label{human_tn_var}
\end{figure}

\begin{figure}[htb]
\centering
\includegraphics[width=\textwidth]{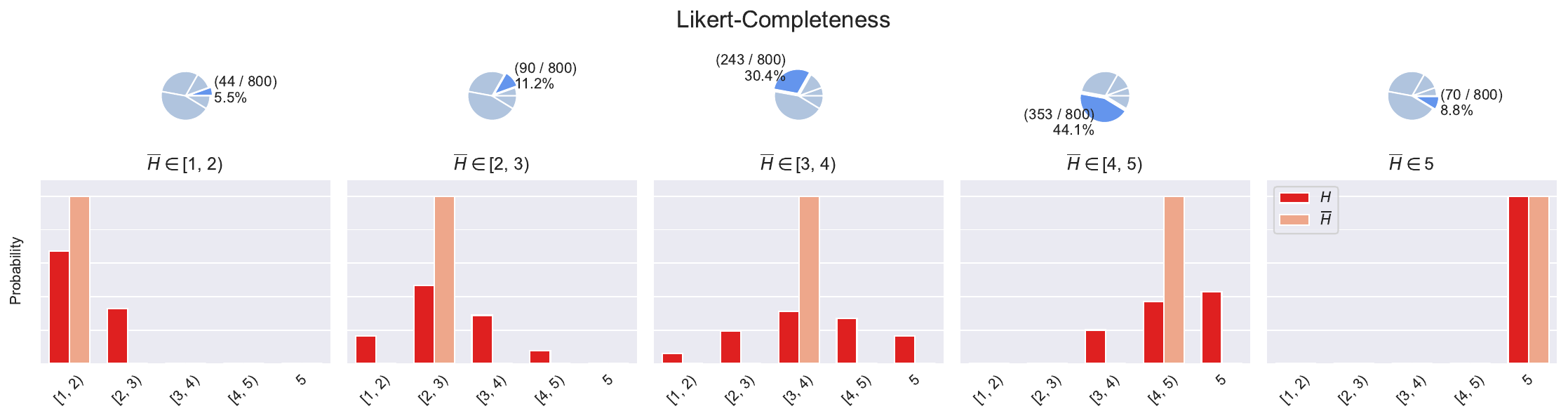}\\
\noindent\rule{0.8\textwidth}{1pt}
\includegraphics[width=\textwidth]{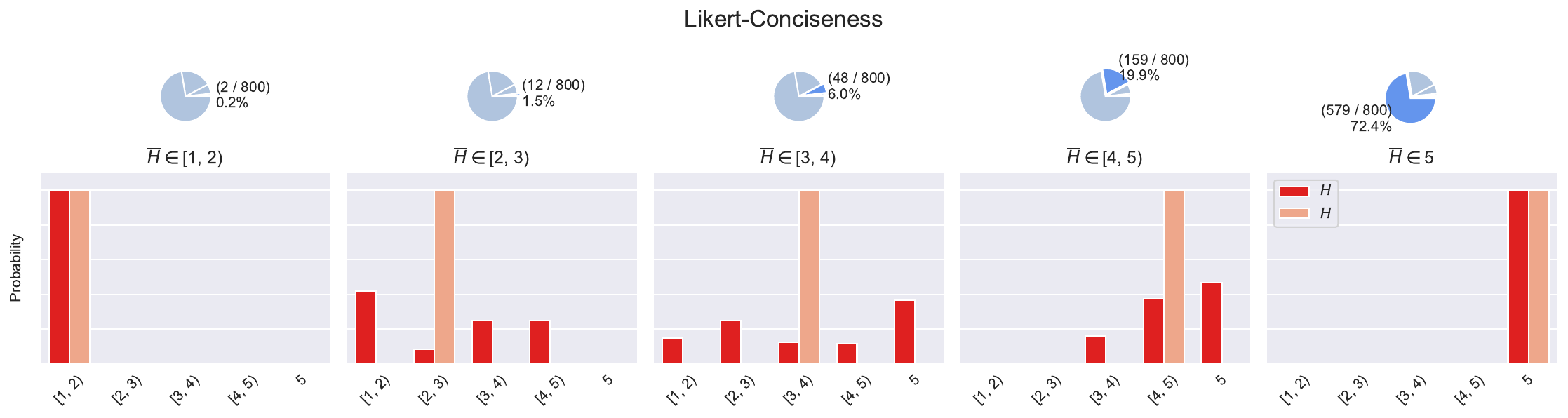}\\
\noindent\rule{0.8\textwidth}{1pt}
\includegraphics[width=\textwidth]{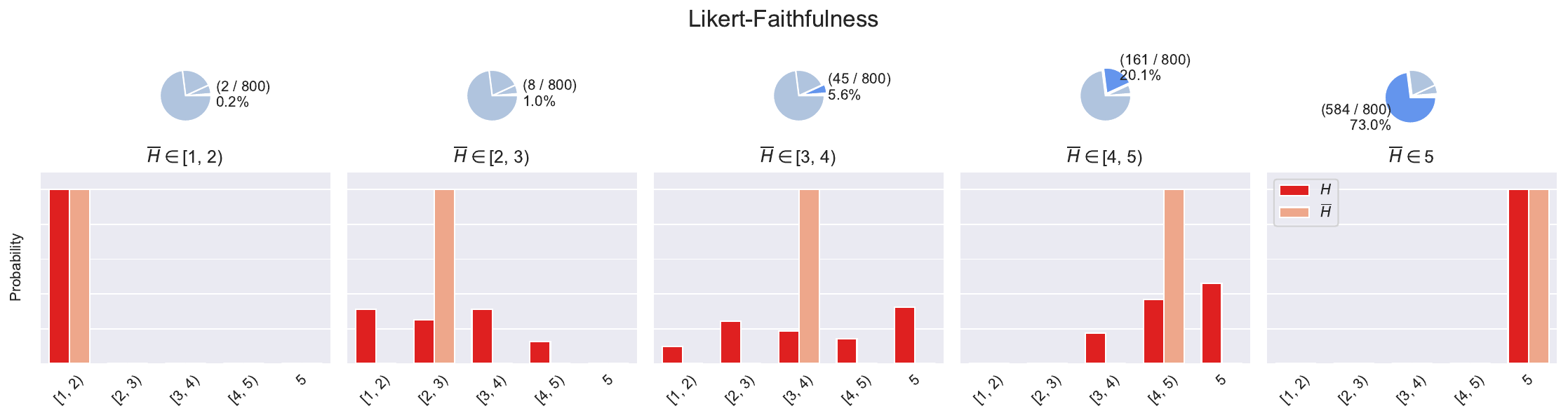}
\caption{Human label distribution for Likert style annotations.}
\label{human_likert_var}
\end{figure}

Figures \ref{human_tn_var} and \ref{human_likert_var} highlight the differences in evaluation methodologies using the visualization method in \citet{elangovan2025beyond}. Despite both methods being expert annotations, \humaneval’s structured rubric-based approach leads to a broader distribution of scores, capturing nuances in note quality. In contrast, Likert-scale ratings tend to cluster, potentially overlooking finer distinctions.

\clearpage
\section{Additional Results}
\subsection{Length of each section in a note by source}

\begin{table}[htb]
\small\centering
\begin{tabular}{ccccc}
\toprule
\textbf{Note Source}     & \textbf{S.}     & \textbf{O.}    & \textbf{A.}   & \textbf{P.}         \\\midrule
\scriptsize Human Notes      & 76 (±57)  & 32 (±21) & 57 (±41) & 29 (±14) \\
\scriptsize Claude 3 Sonnet  & 73 (±23)  & 41 (±10) & 64 (±13) & 71 (±12) \\
\scriptsize Claude 3 Haiku   & 97 (±25)  & 46 (±11) & 77 (±16) & 94 (±22) \\
\scriptsize Llama 3.1 (70B)  & 65 (±15)  & 37 (±13) & 61 (±11) & 75 (±11) \\
\scriptsize Llama 3.1 (8B)   & 94 (±25)  & 56 (±13) & 77 (±17) & 82 (±15) \\
\scriptsize Mistral Large V2 & 88 (±23)  & 51 (±9)  & 65 (±12) & 74 (±11) \\
\scriptsize Mistral (7B)     & 86 (±25)  & 51 (±10) & 66 (±12) & 75 (±11) \\
\bottomrule
\end{tabular}
\caption{Number of words (and standard deviation) in each section.}\label{tab:note_length}
\end{table}

\subsection{Traditional reference-based metrics}
Appendix Table~\ref{tab:reference_based} shows the results for traditional reference-based metrics. Notably, all values look similar, making these n-gram-based metrics insufficient to distinguish LLM performance and provide any meaningful signals for note generation quality. The primary reason is all notes follow a similar structure, with the same section names and fairly standard sentence structure, such as ``The client reports/appears ...''. This structural similarity dominates the n-gram-based metric computation. Therefore, they fail to detect the nuances. 

\begin{table}[htb]
\centering\small
\begin{tabular}{ccccc}
\toprule
\textbf{LLM}     & \textbf{R1-F} & \textbf{R2-F} & \textbf{RL-F} & \textbf{BERT.} \\\midrule
Claude 3 Sonnet  & 39.8          & 10.1          & 20.0          & 87.9               \\
Claude 3 Haiku   & 40.7          & 10.9          & 20.3          & 87.9               \\
Llama 3.1 (70B)  & 41.1          & 10.6          & 20.5          & 88.1               \\
Llama 3.1 (8B)   & 39.4          & 10.4          & 20.2          & 87.6               \\
Mistral Large V2 & 40.1          & 10.3          & 19.9          & 87.9               \\
Mistral (7B)     & 39.9          & 9.7           & 19.5          & 87.9               \\\bottomrule
\end{tabular}
\caption{Reference-based evaluation metrics for notes generated by different LLMs, using human notes as a reference. We show F-measure for ROUGE-1/2/L, as well as BERTScore.}\label{tab:reference_based}
\vspace{-2ex}
\end{table}

\subsection{Inter-Annotator Agreement on importance of rubric items}
Table \ref{table:kappa_alpha_comparison} presents inter-annotator agreement scores among five expert annotators regarding the importance of key characteristics in therapy notes. It includes Krippendorff’s alpha ($\alpha$) and Fleiss’ kappa ($\kappa$) for four main sections—Subjective (S), Objective (O), Assessment (A), and Plan (P)—as well as an overall agreement score. The importance of each characteristic was categorized into five levels: Mandatory, Mandatory in certain circumstances, Highly recommended, Good to have, and Optional. The high agreement scores indicate strong reliability in expert judgments, supporting the structured rubric-based evaluation framework.
\begin{table}[htb]
\centering\small
\begin{tabular}{lccccc}
\toprule
 & \textbf{S} & \textbf{O} & \textbf{A} & \textbf{P}  & \textbf{Overall} \\ \midrule
$\alpha$ & 0.76 & 0.68 & 1.00 & 0.77 & 0.73 \\ 
$\kappa$ & 0.63 & 0.67 & 1.00 & 0.61 & 0.68 \\ \bottomrule
\end{tabular}
\caption{Inter-annotator scores among 5 experts on the \textit{importance} of each key characteristic. S/O/A/P stands for four sections. ``Importance'' has 5 levels: Mandatory, Mandatory in certain circumstances, Highly recommended, Good to have, or Optional. $\alpha$: krippendorff's $\alpha$; $\kappa$: Fleiss' $\kappa$.}\vspace{-2ex}
\label{table:kappa_alpha_comparison}
\end{table}

\newpage

\subsection{ Characteristics of Therapist-Written and LLM-Generated Notes}

Table \ref{rab:rubric_detail} compares the presence of key rubric-based characteristics across therapist-written and LLM-generated notes. It highlights specific rubric items where LLM-generated notes exhibit significantly higher coverage ($20\%$ or more) than human-written notes. 

\begin{table}[tb]
\centering\small
\begin{tabular}{lccc}
\toprule
\textbf{Rubric}     & \textbf{Human} & \textbf{Llama} & \textbf{Mistral} \\\midrule
\textbf{Subjective} &       &       &         \\
chief-complaint     & 78\%  & 75\%  & 78\%    \\
symptoms            & \myhl{56\%}  & 87\%  & 90\%    \\
history             & 59\%  & 56\%  & 59\%    \\
goals               & 33\%  & 40\%  & 42\%    \\
homework            & 1\%   & 1\%   & 3\%     \\
quotes              & 23\%  & 17\%  & 15\%    \\
\midrule
\textbf{Objective}  &       &       &         \\
observed-behavior   & \myhl{53\%}  & 96\%  & 98\%    \\
mental-status       & \myhl{22\%}  & 73\%  & 88\%    \\
assessment-tools    & 10\%  & 5\%   & 7\%     \\
therapy-activities  & 12\%  & 4\%   & 4\%     \\
interventions       & 12\%  & 2\%   & 1\%     \\
\midrule
\textbf{Assessment} &       &       &         \\
diagnosis           & 8\%   & 22\%  & 13\%    \\
triggers            & \myhl{19\%}  & 40\%  & 24\%    \\
progress            & 24\%  & 38\%  & 34\%    \\
analysis            & \myhl{72\%}  & 97\%  & 92\%    \\
response            & 39\%  & 30\%  & 32\%    \\
overall-progress    & 8\%   & 11\%  & 11\%    \\
goals               & 4\%   & 4\%   & 3\%     \\
stages              & 41\%  & 31\%  & 34\%    \\
\midrule
\textbf{Plan}        &       &       &         \\
future-interventions & \myhl{39\%}  & 83\%  & 75\%    \\
follow-up            & 31\%  & 45\%  & 41\%    \\
adjustment           & 2\%   & 9\%   & 7\%     \\
homework             & 33\%  & 33\%  & 26\%    \\
\bottomrule
\end{tabular}
\caption{Characteristics of therapist-written and LLM-generated notes. We highlight rubric items where human is over 20\% lower than the best LLM.}\label{rab:rubric_detail}
\end{table}
\newpage
\subsection{Example human and LLM sourced note for the same conversation}

\begin{table}[htb]
\centering\scriptsize
\begin{tabular}{p{.94\columnwidth}}
\toprule
\textbf{Therapist (Human) written note}\\
\textbf{Subjective}: Client is a male who is presenting to therapy for smoking. \myhl{Client reports his wife quit smoking around six months ago due to being on a health kick.} Client reports he feels pressured to quit smoking by his wife. Client reports his wife is worried about his health and continuously saves articles about the negative effects of smoking and secondhand smoke. \myhl{In addition, he feels pressure as his children feel negatively about his smoking as well.} Client stated "we used to go out and have a cigarette together at the table. Now I have to go outside in this weather. " Client reports feeling upset due to being forced to change his lifestyle habits. Client also feels torn as he enjoys smoking but does not want to lose his family over it. \myhl{Client inquired about his wife possibly attending his next session.}\\
\textbf{Objective}: Client expressed frustration, anger, and reports feeling isolated due to his smoking. \myhl{Client reports his smoking is impacting his family life and his physical health.}\\
\textbf{Assessment}: Client appears to be in the contemplation stage of change. Although client is upset regarding his cigarette smoking, he appears to see the consequences of his smoking. \myhl{In addition, he also appears to have an understanding of consequences regarding losing his family if he continued to smoke.}\\
\textbf{Plan}: Client will follow up with wife to see her schedule in the hopes that she can attend next session with him. Client will also work on reducing his cigarette use. \\\midrule
\textbf{Note generated by Llama 3.1 (70B)}\\
\textbf{Subjective}: Client reports being forced to attend therapy by his wife due to her concerns about his smoking. He expresses feelings of anger, frustration, and resentment towards his wife and children, who he feels are ganging up on him. He also mentions concerns about the impact of his smoking on his family, particularly his wife. He does not see his smoking as a major problem, but rather as a habit he has had all his life.\\
\textbf{Objective}: Client appears resistant and defensive, with a tone of frustration and anger. He maintains a somewhat confrontational demeanor, but also shows signs of vulnerability and concern for his relationship with his wife.\\
\textbf{assessment}: Based on the client's report and observable behavior, it appears that he is struggling with feelings of resentment and anger towards his wife and children, who he feels are pressuring him to quit smoking. He also seems to be motivated by a desire to maintain his relationship with his wife and avoid conflict. It is likely that he is experiencing some ambivalence about quitting smoking, and may benefit from a gradual approach to cessation.\\
\textbf{Plan}: Short-term goals include exploring the client's motivations and concerns about quitting smoking, and developing a gradual plan for cessation. It may be helpful to involve the client's wife in future sessions to address her concerns and provide a unified approach to supporting the client's quit attempt. \myhl{Long-term goals include reducing the client's symptoms of anger and frustration, improving his relationship with his wife and children, and increasing his overall well-being.}\\\bottomrule
\end{tabular}
\caption{Visualized sentences that are considered not concise in human and Llama notes.}\label{tab:concise_example}
\end{table}

\subsection{Section-wise evaluation scores}

\begin{table*}[htb]
\centering\small
\begin{tabular}{cccccccc}
\toprule
                    & \textbf{}                 & \multicolumn{2}{c}{\textbf{Completeness}} & \multicolumn{2}{c}{\textbf{Conciseness}} & \multicolumn{2}{c}{\textbf{Faithfulness}} \\\cmidrule(lr){3-4} \cmidrule(lr){5-6} \cmidrule(lr){7-8}
\textbf{Section}    & \textbf{Note}             & \textbf{\humaneval}     & \textbf{Likert}     & \textbf{\humaneval}     & \textbf{Likert}    & \textbf{\humaneval}     & \textbf{Likert} \\\midrule
Subjective  & Human            & 41.7 (±22.8)    & 3.28 (±1.11)   & 84.7 (±20.8)   & 4.49 (±0.72)   & 92.0 (±15.0)    & 4.64 (±0.67)   \\
            & Llama 3.1 70B    & 46.0 (±12.4)    & 3.86 (±0.74)   & 90.8 (±17.8)   & 4.81 (±0.35)   & 95.0 (±10.9)    & 4.66 (±0.52)   \\
            & Mistral Large V2 & 47.8 (±13.6)    & 4.14 (±0.61)   & 88.7 (±15.4)   & 4.91 (±0.24)   & 97.9 (±5.7)     & 4.87 (±0.40)   \\\cmidrule(lr){1-2}
objective   & Human            & 21.8 (±18.3)    & 2.51 (±1.06)   & 65.9 (±29.9)   & 4.10 (±0.86)   & 85.1 (±23.2)    & 4.40 (±0.74)   \\
            & Llama 3.1 70B    & 36.0 (±8.8)     & 3.56 (±0.87)   & 81.8 (±27.0)   & 4.82 (±0.36)   & 49.0 (±30.0)    & 4.75 (±0.39)   \\
            & Mistral Large V2 & 39.6 (±7.8)     & 3.95 (±0.64)   & 89.0 (±14.7)   & 4.90 (±0.36)   & 60.4 (±28.9)    & 4.94 (±0.26)   \\\cmidrule(lr){1-2}
Assessment  & Human            & 26.9 (±16.1)    & 2.94 (±1.02)   & 83.0 (±23.8)   & 4.35 (±0.81)   & 85.4 (±22.9)    & 4.57 (±0.62)   \\
            & Llama 3.1 70B    & 34.1 (±10.6)    & 3.72 (±0.71)   & 94.7 (±12.2)   & 4.82 (±0.37)   & 80.9 (±22.7)    & 4.70 (±0.52)   \\
            & Mistral Large V2 & 30.4 (±9.9)     & 3.97 (±0.68)   & 95.5 (±11.8)   & 4.80 (±0.44)   & 84.8 (±21.4)    & 4.90 (±0.35)   \\\cmidrule(lr){1-2}
Plan        & Human            & 26.2 (±19.9)    & 2.67 (±1.03)   & 68.4 (±35.9)   & 4.17 (±1.11)   & 78.2 (±33.1)    & 4.13 (±1.08)   \\
            & Llama 3.1 70B    & 42.5 (±19.4)    & 4.05 (±0.76)   & 72.9 (±25.2)   & 4.87 (±0.33)   & 46.6 (±34.2)    & 4.61 (±0.55)   \\
            & Mistral Large V2 & 37.2 (±19.3)    & 3.97 (±0.85)   & 94.4 (±10.1)   & 4.89 (±0.32)   & 43.8 (±34.4)    & 4.88 (±0.33)   \\\bottomrule
\end{tabular}
\caption{Human evaluation results by sections using \humaneval and Likert-style human evaluations.}\label{tab:human_eval_section}
\end{table*}

\newpage
\subsection{Automatic evaluation scores for different note sources and evaluators}

\begin{table*}[htb]
\small\centering
\begin{tabular}{c|ccccccc}
\toprule
\multicolumn{2}{c}{} & \multicolumn{2}{c}{\textbf{Completeness}} & \multicolumn{2}{c}{\textbf{Conciseness}} & \multicolumn{2}{c}{\textbf{Faithfulness}} \\\cmidrule(lr){3-4}\cmidrule(lr){5-6}\cmidrule(lr){7-8}
\textbf{Evaluator} & \textbf{Note Source}                & \textbf{\autoeval}           & \textbf{Likert}         & \textbf{\autoeval}          & \textbf{Likert}         & \textbf{\autoeval}           & \textbf{Likert}         \\\midrule
Mistral  & Human            & 15.0 (±9.1) & 2.23 (±0.27) & 73.7 (±15.1) & 3.65 (±0.53) & 73.2 (±14.9) & 4.64 (±0.39) \\
Large V2 & Claude 3 Sonnet  & 21.7 (±6.5) & 2.67 (±0.32) & 93.6 (±7.8)  & 4.00 (±0.23) & 74.0 (±10.1) & 4.99 (±0.05) \\
         & Claude 3 Haiku   & 21.7 (±6.6) & 2.84 (±0.33) & 94.4 (±6.8)  & 3.92 (±0.22) & 69.9 (±10.1) & 4.92 (±0.21) \\
         & Llama 3.1 (70B)  & 21.0 (±5.4) & 2.53 (±0.25) & 92.3 (±7.7)  & 3.73 (±0.32) & 70.2 (±11.5) & 4.95 (±0.17) \\
         & Llama 3.1 (8B)   & 22.0 (±6.9) & 2.64 (±0.29) & 91.3 (±9.0)  & 3.56 (±0.27) & 69.0 (±11.6) & 4.64 (±0.43) \\
         & Mistral Large V2 & 23.1 (±6.5) & 2.92 (±0.35) & 92.8 (±7.2)  & 3.97 (±0.21) & 75.8 (±8.8)  & 4.99 (±0.05) \\
         & Mistral 7B       & 21.4 (±6.6) & 3.00 (±0.34) & 90.3 (±6.4)  & 4.03 (±0.20) & 75.2 (±9.5)  & 5.00 (±0.00) \\\bottomrule
\end{tabular}
\caption{\papername and Likert-style automatic evaluation. We show the results using Mistral Large V2 as the evaluator. Note that the \papername faithfulness is not LLM-based metric, instead it uses AlignScore.}\label{tab:reference_free_mistral}
\end{table*}

\begin{table*}[htb]
\small\centering
\begin{tabular}{c|ccccccc}
\toprule
\multicolumn{2}{c}{} & \multicolumn{2}{c}{\textbf{Completeness}} & \multicolumn{2}{c}{\textbf{Conciseness}} & \multicolumn{2}{c}{\textbf{Faithfulness}} \\\cmidrule(lr){3-4}\cmidrule(lr){5-6}\cmidrule(lr){7-8}
\textbf{Evaluator} & \textbf{Note Source}                & \textbf{\autoeval}           & \textbf{Likert}         & \textbf{\autoeval}          & \textbf{Likert}         & \textbf{\autoeval}           & \textbf{Likert}         \\\midrule

Llama 3.1 & Human            & 19.7 (±11.1) & 1.77 (±0.33) & 74.8 (±15.3) & 4.68 (±0.40) & 73.2 (±14.9) & 4.63 (±0.50) \\
(70B)     & Claude 3 Sonnet  & 25.0 (±7.2)  & 2.25 (±0.33) & 92.9 (±8.4)  & 4.93 (±0.13) & 74.0 (±10.1) & 5.00 (±0.00) \\
          & Claude 3 Haiku   & 26.9 (±7.0)  & 2.56 (±0.38) & 93.4 (±7.3)  & 4.93 (±0.12) & 69.9 (±10.1) & 4.93 (±0.24) \\
          & Llama 3.1 (70B)  & 24.3 (±6.5)  & 2.19 (±0.28) & 92.3 (±6.8)  & 4.86 (±0.22) & 70.2 (±11.5) & 4.91 (±0.19) \\
          & Llama 3.1 (8B)   & 25.6 (±7.8)  & 2.38 (±0.35) & 92.0 (±8.5)  & 4.63 (±0.46) & 69.0 (±11.6) & 4.67 (±0.46) \\
          & Mistral Large V2 & 28.0 (±7.3)  & 2.46 (±0.38) & 92.8 (±5.5)  & 4.92 (±0.15) & 75.8 (±8.8)  & 4.99 (±0.06) \\
          & Mistral 7B       & 27.8 (±6.7)  & 2.65 (±0.45) & 91.2 (±6.2)  & 4.93 (±0.14) & 75.2 (±9.5)  & 4.98 (±0.09) \\
\bottomrule
\end{tabular}
\caption{\papername and Likert-style automatic evaluation. We show the results using Llama 3.1 (70B) as the evaluator. Note that the \papername faithfulness is not LLM-based metric, instead it uses AlignScore.}\label{tab:reference_free_mistral}
\end{table*}

\section{Workflow Integration Proposal} 
\label{subsec:intergration-proposal}

Below, we outline detailed integration steps for embedding the \papername framework within clinical workflows:

\begin{enumerate}
    \item \textbf{Session Completion:} Therapists conduct standard therapy sessions, optionally recording or leveraging speech-to-text tools integrated with the Electronic Health Record (EHR). We propose using HIPAA-certified tools for this task to ensure client privacy.
    
    \item \textbf{Note Creation:} After completing a session, therapists either write a note from scratch or receive an initial AI-generated SOAP note draft, which they review and edit in the EHR interface. For AI-generated notes, therapists review and manually edit auto-generated drafts within the EHR interface, making necessary adjustments for accuracy and clinical appropriateness. These notes can be in the EHR provider's preferred format.
        
    \item \textbf{\autoeval Quality Assessment:} The \autoeval framework evaluates the edited note in real-time within the EHR, scoring completeness, conciseness, and faithfulness, while providing rubric-aligned actionable feedback.
    
    \item \textbf{Verification and Final Submission:} Therapists review the \autoeval quality scorecard and address highlighted concerns before formally submitting notes to the EHR, maintaining final responsibility and clinical oversight.
    \end{enumerate}
\end{document}